\documentclass{article}

\usepackage{arxiv}
\usepackage[utf8]{inputenc} 
\usepackage[T1]{fontenc}    
\usepackage{hyperref}       
\usepackage{url}            
\usepackage{booktabs}       
\usepackage{amsfonts}       
\usepackage{nicefrac}       
\usepackage{microtype}      
\usepackage{graphicx}
\usepackage[square,sort,comma,numbers]{natbib}
\usepackage{doi}
\usepackage{amsmath}
\usepackage{stmaryrd}
\usepackage{ifthen}
\usepackage{algorithm, algorithmic}
\usepackage{longtable}
\usepackage{caption}
\usepackage{subcaption}

\title{NeuroPrim: An Attention-based Model for Solving NP-hard Spanning Tree Problems}

\author{ 
    {Yuchen Shi} \\
    Department of Mathematical Sciences\\
    University of Chinese Academy of Sciences\\
	Beijing {\rm 100049}, China \\
	\texttt{shiyuchen20@mails.ucas.ac.cn}
	\And
	{Congying Han} \\
    Department of Mathematical Sciences\\
    University of Chinese Academy of Sciences\\
	Beijing {\rm 100049}, China \\
	\texttt{hancy@ucas.ac.cn}
    \And
    {Tiande Guo}\thanks{Corresponding author} \\
    Department of Mathematical Sciences\\
    University of Chinese Academy of Sciences\\
	Beijing {\rm 100049}, China \\
	\texttt{tdguo@ucas.ac.cn}
}

\renewcommand{\headeright}{}
\renewcommand{\undertitle}{}
\renewcommand{\shorttitle}{Accepted for publication in SCIENCE CHINA Mathematics}

\begin{document}
\maketitle

\begin{abstract}
	Spanning tree problems with specialized constraints can be difficult to solve in real-world scenarios, often requiring intricate algorithmic design and exponential time. Recently, there has been growing interest in end-to-end deep neural networks for solving routing problems. However, such methods typically produce sequences of vertices, which makes it difficult to apply them to general combinatorial optimization problems where the solution set consists of edges, as in various spanning tree problems. In this paper, we propose NeuroPrim, a novel framework for solving various spanning tree problems by defining a Markov Decision Process (MDP) for general combinatorial optimization problems on graphs. Our approach reduces the action and state space using Prim's algorithm and trains the resulting model using REINFORCE. We apply our framework to three difficult problems on Euclidean space: the Degree-constrained Minimum Spanning Tree (DCMST) problem, the Minimum Routing Cost Spanning Tree (MRCST) problem, and the Steiner Tree Problem in graphs (STP). Experimental results on literature instances demonstrate that our model outperforms strong heuristics and achieves small optimality gaps of up to 250 vertices. Additionally, we find that our model has strong generalization ability, with no significant degradation observed on problem instances as large as 1000. Our results suggest that our framework can be effective for solving a wide range of combinatorial optimization problems beyond spanning tree problems.
\end{abstract}


\section{Introduction}

Network design problems are ubiquitous and arise in various practical applications. For instance, while designing a road network, one needs to ensure that the number of intersections does not exceed a certain limit, giving rise to a Degree-constrained Minimum Spanning Tree (DCMST) problem. Similarly, designing a communication network requires minimizing the average cost of communication between two users, which can be formulated as a Minimum Routing Cost Spanning Tree (MRCST) problem. In addition, minimizing the cost of opening facilities and communication network layouts while maintaining interconnectivity of facilities is critical for telecommunication network infrastructure. This problem can be modeled as a Steiner Tree Problem in graphs (STP). Although these problems all involve finding spanning trees in graphs, they have different constraints and objective functions, requiring heuristic algorithms tailored to their specific problem characteristics, without a uniform solution framework. Moreover, unlike the Minimum Spanning Tree (MST) problem, which can be solved in polynomial time by exact algorithms such as Prim's algorithm \cite{primShortestConnectionNetworks1957}, finding optimal solutions for these problems requires computation time exponential in problem size, making it impractical for real-world applications.

In contrast to conventional algorithms that are designed based on expert knowledge, end-to-end models leveraging deep neural networks have achieved notable success in tackling a wide range of combinatorial optimization problems \cite{guoSolvingCombinatorialProblems2019, guoMachineLearningMethods2019}. However, most methods treat the problem as permutating a sequence of vertices, e.g., allocating resources in the knapsack problem \cite{dingHighGeneralizationPerformance2021}, finding the shortest travel route in routing problems \cite{vinyalsPointerNetworks2017, belloNeuralCombinatorialOptimization2017,nazariReinforcementLearningSolving2018, koolAttentionLearnSolve2019, wangLearningTraverseGraphs2021, wangGameTheoreticApproachImproving2022}, approximating minimum vertex cover and  maximum cut on graphs \cite{daiLearningCombinatorialOptimization2018}, learning elimination order in tree decomposition \cite{khakhulinLearningEliminationOrdering2020}, locating facilities in the p-median problem \cite{wangSolvingUncapacitatedPMedian2022}, and determining Steiner points in STP \cite{ahmedComputingSteinerTrees2021}. For problems that require decision-making on edge sets such as MST, due to the need to consider the connectivity and acyclicity of the solution, there hardly exists methods that deal with the edge generation tasks directly. \cite{droriLearningSolveCombinatorial2020} proposed to first transform edges into vertices by constructing line graphs, then use an end-to-end approach to tackle various graph problems, including MST. This method, however, is not suitable for dense graphs, as their line graphs are quite large. \cite{duVulcanSolvingSteiner2021} presented the use of Graph Neural Networks and Reinforcement Learning (RL) to solve STP, whose decoder, after outputting a vertex, selects the edge with the smallest weight connected to that vertex in the partial solution. However, this method cannot be applied to problems where the objective function is not the sum of the weights of all edges.

This paper proposes NeuroPrim, a novel framework that utilizes the successful Transformer model to solve three challenging spanning tree problems in Euclidean space, namely the DCMST, MRCST and STP problems. The algorithm is based on a Markov Decision Process (MDP) for general combinatorial optimization problems on graphs, which significantly reduces the action and state spaces by maintaining the connectivity of partially feasible solutions, similar to Prim's algorithm. The framework is parameterized and trained with problem-related masks. Our goal is not to outperform existing solvers; instead, we demonstrate that our approach can produce approximate solutions to various sizes and variants of spanning tree problems quickly, with minimal modifications required after training. Our approach's effectiveness may provide insights into developing algorithms for problems with complex constraints and few heuristics.

Our work makes several significant contributions, including:
\begin{itemize}
    \item We present a novel Markov Decision Process (MDP) framework for general combinatorial optimization problems on graphs. This framework allows all constructive algorithms for solving these problems to be viewed as greedy rollouts of some policy solved by the model.
    \item We propose NeuroPrim, a unified framework that leverages the MDP to solve problems that require decisions on both vertex and edge sets. Our framework is designed to be easily adaptable, requiring only minimal modifications to masks and rewards to solve different problems.
    \item To the best of our knowledge, we are the first to use a sequence-to-sequence model for solving complex spanning tree problems. We demonstrate the effectiveness of our approach by applying it to three challenging variants of the spanning tree problem on Euclidean space, achieving near-optimal solutions in a short time and outperforming strong heuristics.
\end{itemize}

\section{Preliminaries}
In this section, we describe the problems investigated and the techniques employed. Firstly, we present three variants of the spanning tree problem under investigation. Subsequently, we introduce the networks utilized for policy parameterization.

\subsection{Spanning tree problems}

\begin{figure}[!htbp]
	\centering
	\includegraphics[width=\textwidth, angle=0]{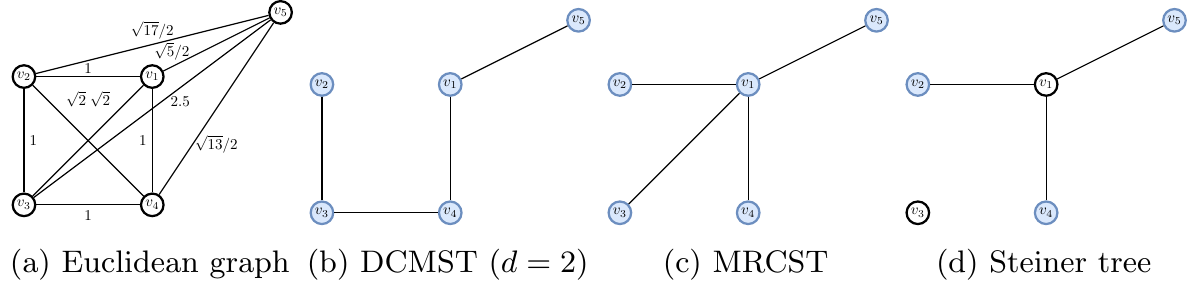}
	\caption{Examples of optimal solutions to different spanning tree problems. All vertices to be spanned (i.e., all vertices in DCMST and MRCST problem and terminals in STP) are colored in blue. The black lines in (a) indicates a feasible edge in graph, and in (b), (c), (d) an edge in the optimal solution. (a) A complete Euclidean graph with five vertices. (b) MST and also DCMST ($d=2$) of the graph with a total weight of $3+\sqrt{5}/2$. (c) MRCST with overall pairwise distance of $8+4\sqrt{2}+2\sqrt{5}$. (d) A Steiner tree connecting terminals $\{v_2,v_4,v_5\}$ with total weight of $2+\sqrt{5}/2$, where $\{v_1\}$ is the Steiner point.}
	\label{fig:3stps}
\end{figure}

Given an undirected weighted graph $(V(G), E(G), w)$ with $v(G)$ vertices, $e(G)$ edges and $w:E\rightarrow\mathbb{R}$ specifying the weight of each edge, a spanning tree of $G$ is a subgraph that is a tree which includes all vertices of $G$. Among all combinatorial optimization problems with spanning trees, the Minimum Spanning Tree (MST) problem is one of the most typical and well-known one. The problem seeks to find a spanning tree with least total edge weights and can be solved in time $O(e(G)\log v(G))$ by Borůvka's algorithm \cite{boruvkaJisiemProblemuMinimalnim1926}, Prim's algorithm \cite{primShortestConnectionNetworks1957} and Kruskal's algorithm \cite{kruskalShortestSpanningSubtree1956}. However, with additional constraints or modifications to the objective function, it becomes more difficult to solve these problems to optimality. Three variants of the spanning tree problem are considered in this paper. For each variant, an example in Euclidean space is shown in Figure \ref{fig:3stps}.

\subsubsection{Degree-constrained Minimum Spanning Tree problem}

Given an integer $d\ge 2$, the degree-constrained minimum spanning tree (DCMST) problem (also known as the minimum bounded degree spanning tree problem) involves finding a MST of maximum degree at most $d$. The problem was first formulated by \cite{narulaDegreeconstrainedMinimumSpanning1980}. The NP-hardness can be shown by a reduction to the Hamilton Path Problem when $d=2$ \cite{gareyComputersIntractabilityGuide1979}. Several exact \cite{caccettaBranchCutMethod2001, andradeUsingLagrangianDual2006, dacunhaLowerUpperBounds2007, bicalhoBranchandcutandpriceAlgorithmsDegree2016, fagesSalesmanTreeImportance2016}, heuristic \cite{boldonMinimumweightDegreeconstrainedSpanning1996}, metaheuristic \cite{raidlEdgeSetsEffective2003, delbemNodeDepthEncodingEvolutionary2004, buiAntbasedAlgorithmFinding2006,doanEffectiveAntbasedAlgorithm2007, buiImprovedAntBasedAlgorithm2012, singhHybridGeneticAlgorithm2020} algorithms and approximation schemes \cite{furerApproximatingMinimumDegreeSteiner1994, goemansMinimumBoundedDegree2006, singhApproximatingMinimumBounded2015} have been proposed to tackle this problem. In Figure \hyperref[fig:3stps]{1b}, the path $v_2v_3v_4v_1v_5$ is the MST with maximum degree $2$.

\subsubsection{Minimum Routing Cost Spanning Tree problem}

The minimum routing cost spanning tree (MRCST) problem (also known as the optimum distance spanning tree problem) aims to minimizing the sum of pairwise distances between vertices in the tree, i.e., finding a spanning tree of $G$ that minimizes $\sum_{(x,y)\in V(G)\times V(G), x<y}d_H(x, y)$, where $d_H(x, y)$ is the total weights of the unique path between vertices $x,y\in V(G)$ in the tree $H$. This problem is first stated by \cite{scottOptimalNetworkProblem1969} and then its NP-hardness is raised by \cite{johnsonComplexityNetworkDesign1978}. Exact \cite{fischettiExactAlgorithmsMinimum2002,tilkCombinedColumnandrowgenerationOptimal2018, zetinaSolvingOptimumCommunication2019, agarwalNewValidInequalities2019,choqueOptimalCommunicationSpanning2021}, heuristic \cite{camposFastAlgorithmComputing2008, singhNewHeuristicMinimum2008, tanHeuristicApproachSolving2012, sattariVariableNeighborhoodSearch2013, masoneMinimumRoutingCost2019}, metaheuristic \cite{tanGeneticApproachSolving2012, singhArtificialBeeColony2011, sattariMetaheuristicAlgorithmMinimum2015} algorithms and approximation schemes \cite{wongWorstCaseAnalysisNetwork1980, wuPolynomialTimeApproximationScheme2000} have been developed for this problem. In Figure \hyperref[fig:3stps]{1c}, the star graph with internal vertex $v_1$ and leaves $v_2$, $v_3$, $v_4$ and $v_5$ is the MRCST, and its routing cost is 4 times its total weight. (Note that after deleting any edge, the graph becomes two connected components with orders $1$ and $4$, so that the contribution of each edge to the routing cost is $1\cdot 4=4$.)

\subsubsection{Steiner Tree Problem in graphs}

Given a subset of vertices $V^{\prime}\subseteq V(G)$ called terminals, the Steiner tree problem in graphs (STP) asks for a tree in $G$ that spans $V^{\prime}$ with minimum weights, i.e., finding a set $S\subseteq V(G)\setminus V^{\prime}$ that minimize the cost of the MST of $G[V^{\prime}\cup S]$, where $G[V^{\prime}\cup S]$ is the subgraph induced by $V^{\prime}\cup S$ and $S$ is called Steiner points. The problem is NP-hard \cite{karpREDUCIBILITYCOMBINATORIALPROBLEMS1972} and has two polynomially solvable variants: shortest path problem ($|V^{\prime}|=2$) and MST problem ($|V^{\prime}|=v(G)$). Exact \cite{lucenaBranchCutAlgorithm1998, kochSolvingSteinerTree1998, dearagaoDualHeuristicsExact2001,polzinImprovedAlgorithmsSteiner2001}, heuristic \cite{takahashiApproximateSolutionSteiner1980, duinEfficientPathVertex1997, ribeiroHybridGRASPPerturbations2002, uchoaFastLocalSearch2012}, metaheuristic \cite{kapsalisSolvingGraphicalSteiner1993, esbensenComputingNearoptimalSolutions1995, prestiGridEnabledParallel2004, huySolvingGraphicalSteiner2008, ribeiroTabuSearchSteiner2000, bastosReactiveTabuSearch2002, bouchachiaHybridEnsembleApproach2011, quParticleSwarmOptimization2013} algorithms and approximation schemes \cite{aroraPolynomialTimeApproximation1998, chlebikSteinerTreeProblem2008, byrkaImprovedLPbasedApproximation2010} have been devised to address this problem. The most recent breakthrough lies in the 11th DIMACS (the Center for Discrete Mathematics and Theoretical Computer Science) Implementation Challenge and the PACE 2018 Parameterized Algorithms and Computational Experiments Challenge \cite{bonnetPACE2018Parameterized2018}. In Figure \hyperref[fig:3stps]{1d}, the star graph with internal vertex $v_1$ and leaves $v_2$, $v_4$ and $v_5$ is the Steiner tree connecting $\{v_2, v_4,v_5\}$.

\subsection{Transformer network}

The Transformer \cite{vaswaniAttentionAllYou2017} is a neural network architecture that was introduced in 2017 by Vaswani et al. It was designed to address some of the limitations of traditional recurrent neural networks (RNNs) and convolutional neural networks (CNNs) in processing sequential data. The Transformer has become a popular choice for various natural language processing (NLP) tasks, such as language translation, language modeling, and text classification.

\begin{figure}[!htbp]
	\centering
    \includegraphics[width=\textwidth, angle=0]{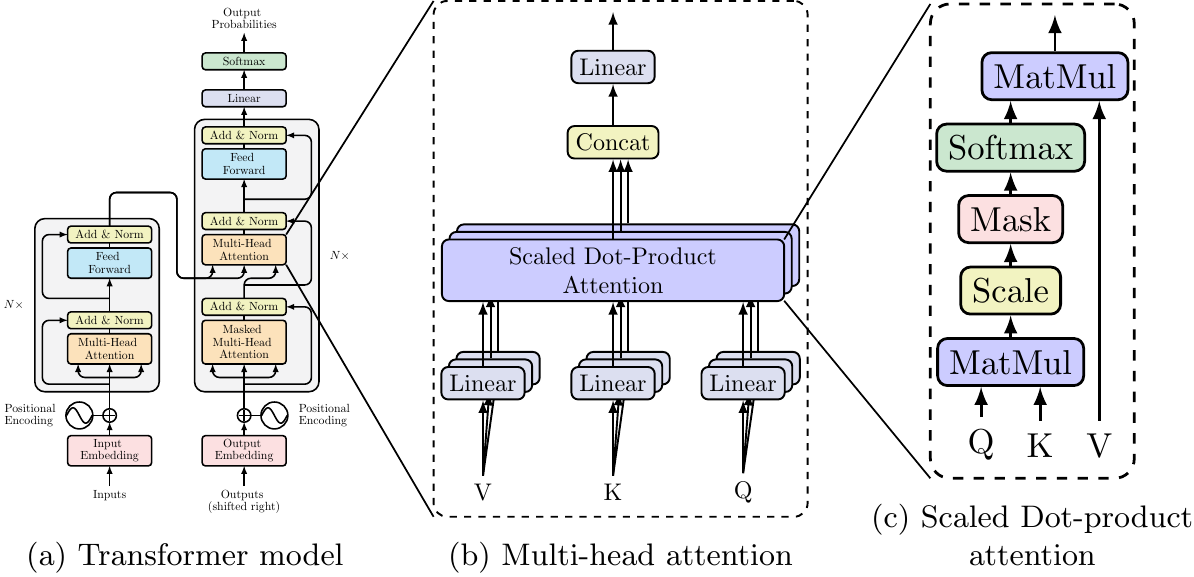}
	\caption{The Transformer model architecture.}
	\label{fig:transformer}
\end{figure}

The Transformer model architecture, as illustrated in Figure \hyperref[fig:transformer]{2a}, consists of two main components: the encoder and the decoder. The encoder takes the input sequence and produces a set of hidden states that represent the input sequence. The decoder takes the output of the encoder and produces the output sequence. Both the encoder and decoder consist of multiple layers of multi-head attention (MHA) and feed-forward networks (FFN). The MHA mechanism is the key component that allows the Transformer to model long-range dependencies in the input sequence.

Self-attention is a mechanism that allows the model to attend to different parts of the input sequence to compute a representation for each part. The self-attention mechanism employs query and key vectors to compute the attention weights, which determine the contribution of each element in the input sequence to the final output. The attention weights are computed by taking the dot product of the query and key vectors, dividing the result by the square root of the dimensionality of the key vectors, and normalizing the scores using the softmax function as depicted in Figure \hyperref[fig:transformer]{2c}. MHA extends self-attention by performing attention over multiple projections of the input sequence. In MHA, the input sequence is projected into multiple subspaces, and attention is performed separately on each subspace. This allows the model to attend to different aspects of the input sequence simultaneously, providing a richer representation of the input sequence. Specifically, the MHA block with residual connection (RC) and batch normalization (BN) can be defined as a parameterized function class $\operatorname{MHA}:\mathbb{R}^{n\times d}\times \mathbb{R}^{m\times d}\times \mathbb{R}^{m\times d}\times \{0,1\}^{n\times m}\rightarrow \mathbb{R}^{n\times d}$ given by $\operatorname{MHA}(Q,K,V,M)=\operatorname{BN}(Q+O)$, where
\begin{equation}
    O=\bigparallel_{h=1}^{H}\Big(\operatorname{softmax}\Big(M\odot(QW^q_h+\boldsymbol{1}_n b^q)(KW^k_h+\boldsymbol{1}_m b^k)^T\slash\sqrt{d}\Big)(VW^v_h+\boldsymbol{1}_m b^v)\Big)W^o_h + \boldsymbol{1}_n b^o,
\end{equation}
$\bigparallel$ is the matrix concatenation operator, $d$ is the embedding dimension of the model, $H$ is the number of parallel attention heads satisfying that $d$ is divisible by $H$, $\odot$ represents the element-wise product, $\boldsymbol{1}_n=[1\ \cdots\ 1]^T\in\mathbb{R}^{n\times 1}$, and $W^q_h, W^k_h,W^v_h\in\mathbb{R}^{d\times(d\slash H)}$, $b^q,b^k,b^v\in\mathbb{R}^{1\times d}$, $W^o\in\mathbb{R}^{d\times d}$, $b^o\in\mathbb{R}^{1\times d}$ and $h\in\{1,\ldots,H\}$ are trainable parameters. The MHA block is shown in Figure \hyperref[fig:transformer]{2b}.

The FFN with RC and BN can be similarly defined as $\operatorname{FFN}:\mathbb{R}^{n\times d}\rightarrow\mathbb{R}^{n\times d}$, giving
\begin{equation}
\operatorname{FFN}(X)=\operatorname{BN}\left(X+\left(\operatorname{ReLU}(XW_1+\boldsymbol{1}_n b_1)W_2+\boldsymbol{1}_n b_2\right)\right),
\end{equation}
where $d_f$ is the dimension of hidden layer, and $W_1\in\mathbb{R}^{d\times d_f}$, $b_1\in\mathbb{R}^{1\times d_f}$, $W_2\in\mathbb{R}^{d_f\times d}$ and $b_2\in\mathbb{R}^{1\times d}$ are trainable parameters.

\section{Method}

Our approach begins by defining an MDP for general combinatorial optimization problems on graphs. We then present the NeuroPrim algorithm and its constituent components for solving three variants of the spanning tree problem (i.e., the DCMST, MRCST, and STP problems) on Euclidean space. The algorithm constructs trees incrementally, one edge at a time, utilizing a parameterized policy, while ensuring tree connectivity as in the Prim's algorithm.

\subsection{MDP formulation}

Many combinatorial optimization problems on graphs can be modeled as an MDP and solved approximately using RL. Formally, such problems can be specified as $(\mathcal{I}, f,\mu)$, where $\mathcal{I}$ is a set of graph instances, $f(G)$ is the finite set of all feasible solutions of $G\in \mathcal{I}$, and $\mu(G, H)$ denotes the problem-related measure of $H\in f(G)$. Taking only minimization problems into consideration, the solution is $\operatornamewithlimits{argmin}_{H\in f(G)}\mu(G,H)$ for $G\in\mathcal{I}$. Accordingly, The MDP $(\mathcal{S}, \mathcal{A}, P_a, R_a)$ can be defined as follows:
\begin{itemize}
	\item $\mathcal{S}=\{s\mid s\subseteq H, H\in f(G), G\in \mathcal{I}\}$ is the state space. The starting state $s_0$ is the null graph $K_0$, and the set of terminal states $\mathcal{T}=\{H\mid H\in f(G), G\in\mathcal{I}\}$.
	\item $\mathcal{A}=V(G)\cup E(G)$ is the action space, and $\mathcal{A}_s=(V(G)\setminus V(s))\cup (E(G)\setminus E(s))$ is the set of actions available from state $s$.
	\item The transition function $P:\mathcal{S}\times\mathcal{A}\rightarrow\Delta\mathcal{S}$ is defined as
	\begin{equation}
		P(s^\prime\mid s,a)=
		\begin{cases}
		1 &\text{if}\ s\in\mathcal{T}\text{, and}\ s^\prime=s\\
		1 &\text{if}\ s\in\mathcal{S}\setminus\mathcal{T}\text{, and}\ s^\prime=s+a\\
		0 &\text{otherwise}
		\end{cases},
	\end{equation}
	where $\Delta\mathcal{S}$ is the space of probability distributions over $\mathcal{S}$ and $s+a$ denotes the disjoint union of $s$ and $a$.
	\item The reward function $r:\mathcal{S}\times\mathcal{A}\times\mathcal{S}\rightarrow\mathbb{R}$ is given by $r(s,a,s^\prime)=-\mu(G,s^\prime)$ if $s\notin\mathcal{T}$ and $s^\prime\in\mathcal{T}$, otherwise $0$.
\end{itemize}

The goal of RL is to learn a policy $\pi:\mathcal{S}\rightarrow\Delta\mathcal{A}$ to maximizes the flat partial return $\bar{G}_{0:T}=\sum_{t=1}^{T}R_t=R_T$, where $R_t$ is the reward received at time step $t$ and $T$ is the final time step to reach any terminal state. Note that the Borůvka's, Prim's and Kruskal's algorithm for solving the MST problem and constructive heuristics for approximating the Traveling Salesman Problem can all be regarded as greedy polices.

\subsection{NeuroPrim algorithm}

Given the prohibitively large size of the action and state spaces defined by the MDP, we introduce the NeuroPrim algorithm. Our approach utilizes techniques inspired by the Prim's algorithm, significantly reducing the state and action spaces. As a result, we can approximate the considered spanning tree problems efficiently using Reinforcement Learning.

We first give the specific forms of $\mathcal{A}_s$, $\mu$ and $T$ in the MDP of these problems. Since the elements in $f(G)$ are all spanning subgraphs of $G$, and taking advantage of the policy optimization with multiple optima (POMO) \cite{kwonPOMOPolicyOptimization2020} training method, the action space at $t=0$ is set as the vertices that must be visited. By ensuring the connectivity of state $s$ as in Prim's algorithm, $\mathcal{A}_s$ can be further reduced to
\begin{equation}
	\mathcal{A}_s=
	\begin{cases}
		\{V(G)\}&\text{if}\ s=K_0\\
		\{xy\mid x\in V(s), y\notin V(s)\} &\text{else}
	\end{cases}.
\end{equation}
For STP, since the non-terminal points will not necessarily be in the optimal solution, we set $\mathcal{A}_{K_0}=V'\subseteq V(G)$.
$\mu(G,H)$ is the cost of feasible solution $H\subseteq G$ for each problem, i.e., $\sum_{xy\in H}w(xy)$ for the DCMST and STP problems, and $\sum_{(x,y)\in V(G)\times V(G), x<y}d_H(x, y)$ for the MRCST problem. $T$ is exactly $v(G)-1$ in DCMST and MRCST problems, and lies between $|V'|$ and $v(G)$ in STP, as determined by whether all terminals are spanned.

\begin{figure}[!htbp]
    \includegraphics[width=\textwidth, angle=0]{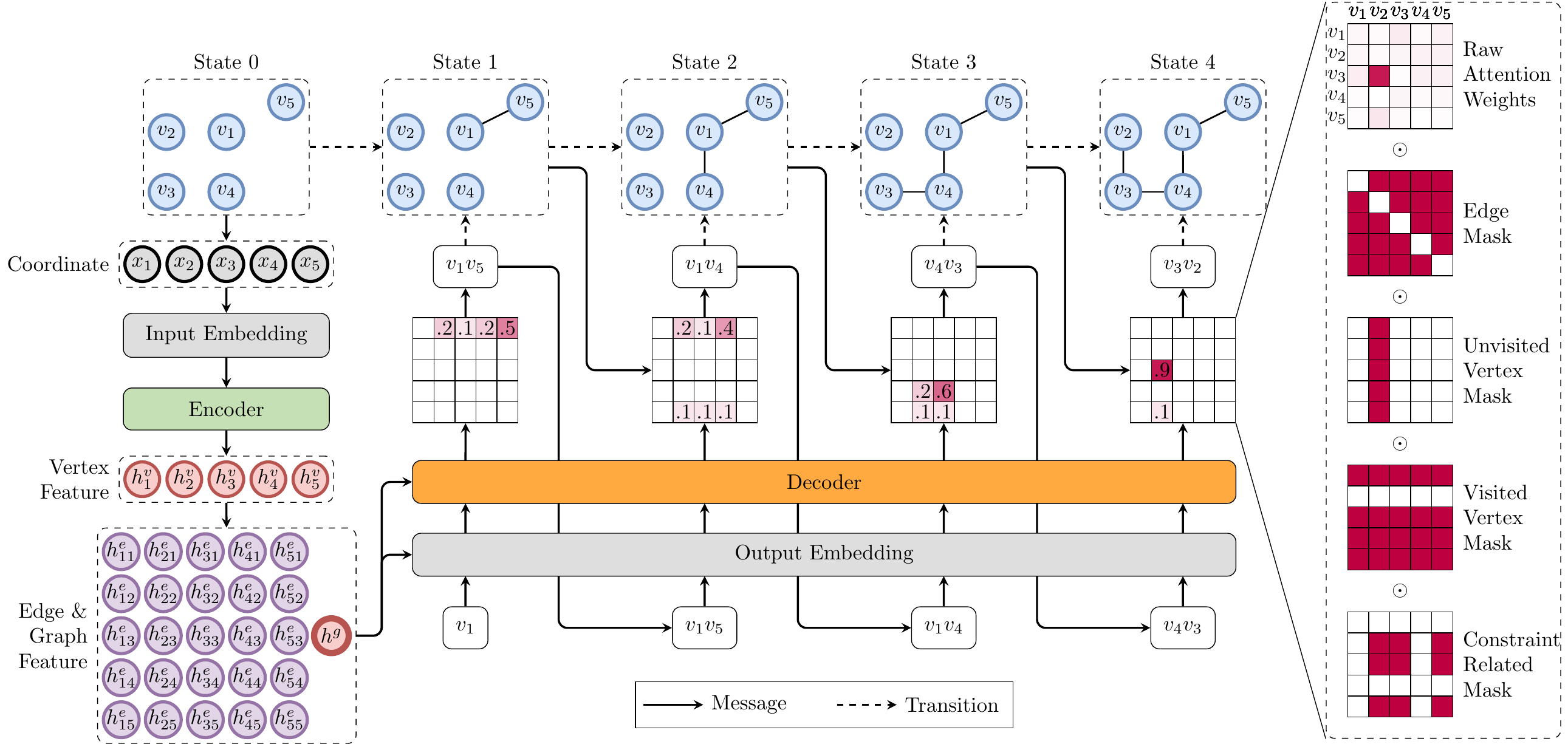}
	\caption{The NeuroPrim algorithm solves the DCMST ($d=2$) problem by generating a sequence of actions that form the optimal solution ($v_2,v_3,v_4,v_1,v_5$). The dashed arrows illustrate the state transitions in the Markov Decision Process (MDP), while the solid arrows indicate the flow of parameterized state and action messages. The blue vertices and edges represent the current state, starting from the initial state $K_0$. The encoder processes the graph coordinates to obtain vertex, edge, and graph embeddings. At each time step, the decoder takes the previous actions and graph features as inputs, and produces masked attention weights to determine the next action. The mask is computed through a Hadamard product of four masks, with the purple element indicating the edge index that can be selected.
    }
	\label{fig:neuroprim}
\end{figure}

Drawing inspiration from \cite{bressonTransformerNetworkTraveling2021}, the policy is parameterized by vanilla Transformer model \cite{vaswaniAttentionAllYou2017} as $\pi_{\boldsymbol{\theta}}(A_t\mid G, S_t)$, which takes as input the encoder embedding of graph $G$ and the concatenated decoder output representing $S_t$, and outputs the probability of available actions from state $S_t$. The goal is to train $\pi_{\boldsymbol{\theta}}$ to maximize the expected flat partial return $J(\boldsymbol{\theta})=\mathbb{E}_{\pi_{\boldsymbol{\theta}}}[R_T]$, where $\mathbb{E}_{\pi_{\boldsymbol{\theta}}}[\cdot]$ is taken over all trajectories under $\pi_{\boldsymbol{\theta}}$, each with a generation probability of $\prod_{t=0}^{T-1} \pi_{\boldsymbol{\theta}}(A_t\mid G,S_t)$. The model is trained by REINFORCE using the POMO baseline with sampling. Algorithm \ref{alg:neuroprim} and Figure \ref{fig:neuroprim} illustrate the proposed NeuroPrim algorithm and a corresponding example, respectively.

\begin{algorithm}
  \caption{NeuroPrim algorithm}\label{alg:neuroprim}
  \begin{algorithmic}[1]
    \renewcommand{\algorithmicrequire}{\textbf{Input:}}
    \REQUIRE{graph $G$, starting vertex $v_0\in V(G)$}
    \STATE{$s\leftarrow G[v_0]$}
    \WHILE{$s\notin \mathcal{T}$}
        \STATE{Calculate $\mathcal{A}_s$ as Prim's algorithm}
        \STATE{Choose $a\in\mathcal{A}_s$ from $\pi_{\boldsymbol{\theta}}(\cdot\mid G,s)$}
        \STATE{$s\leftarrow s+a$}
    \ENDWHILE
    \RETURN{$s$}
  \end{algorithmic}
\end{algorithm}

\subsection{Feature aggregation}

The algorithm requires aggregated vertex, graph and edge features as input, which is obtained by a purely self-attentive based encoder.

The encoder first computes the initial vertex embeddings $H^v_0$ through a trainable linear projection with parameters $W^x\in \mathbb{R}^{2\times d_v}$ and $b^x\in\mathbb{R}^{1\times d_v}$, i.e., $H^v_0=XW^x+\boldsymbol{1}_{v(G)}b^x$, where  $X$ is the 2-dimensional coordinates of instance $G$ and $d_v$ is the dimension of vector features. The vertex features $h^v_i$ are then calculated by
\begin{align}
    H^v_{l}&=\operatorname{FFN}\left(\operatorname{MHA}\left(H^v_{l-1},H^v_{l-1},H^v_{l-1},J_{v(G)}\right)\right),\\
    [h^v_1\ \cdots\ h^v_{v(G)}]&=(H^v_{L^{\text{enc}}})^T,
\end{align}
where $l\in\{1,\ldots L^{\text{enc}}\}$ is the encoder layer index. Vertices of different classes (e.g., terminal and non-terminal vertices) do not share parameters.

The graph feature $h^g$ is the aggregation of all vertex features, i.e., $h^g={W^g}\left(\sum_{i=1}^{v(G)}h^v_i\slash v(G)\right)+{b^g}$, where $W^g\in\mathbb{R}^{d_e\times d_v}$ and $b^g\in\mathbb{R}^{d_e}$ are trainable parameters.

One popular approach to obtain edge features is to use a function that takes the feature vectors of the two nodes connected by an edge $(i,j)$ as input and outputs a corresponding edge feature $h^e_{ij}$. One such function is given by the equation:
\begin{equation}
    h^{e}_{ij} = \left[\begin{smallmatrix}h^v_i - h^v_j\\ \max(h^v_i, h^v_j)\end{smallmatrix}\right],
\end{equation}
where $h^v_i$ and $h^v_j$ are the feature vectors of nodes $i$ and $j$, respectively. The first component of the edge feature vector corresponds to the difference between the feature vectors of the two nodes, which characterizes the relative discrepancy or distance their attributes and is informative for capturing the directionality of the relationship between them. The second component is the maximum of the feature vectors of the two nodes, which represents the shared information between the two nodes and reflects the similarity or dissimilarity between the features of the two nodes. The edge features are subsequently linearly transformed to $H^e$ for later use in decoding:
\begin{equation}
    H^{e} = [h^{e}_{11}\ \cdots\ h^{e}_{1n}\ \cdots\ h^{e}_{n1}\ \cdots\ h^{e}_{nn}]^TW^e+\boldsymbol{1}_{e(G)}b^e,
\end{equation}
where $W^e\in\mathbb{R}^{2d_v\times d_e}$ and $b^e\in\mathbb{R}^{1\times d_e}$ are trainable parameters, and $d_e$ is the dimension of edge features.

\subsection{Policy parameterization}

The parameterization of the policy relies on an autoregressive decoder, which uses the information from previous time steps to generate the value of the current time step. Let $A_{0:{t-1}}=(v_0,u_1v_1,\ldots,u_{t-1}v_{t-1})$ denotes the action list and also the partial solution before time step $t$, where $v$ is a randomly selected vertices to be spanned. We take the graph feature together with the feature representation of $A_{0:t-1}$ as input to the decoder at time $t$:
\begin{equation}
    h_{t,0}=\operatorname{MaxPool}\left(\left[\begin{smallmatrix}h^g\ \\h^e_{u_{t-1}v_{t-1}}\end{smallmatrix}\right]\right),
\end{equation}
where $u_0=v_0$ and $\operatorname{MaxPool}$ is a downsampling operator with kernel size 2.

The hidden state $h_{t,L^{\text{dec}}}$ is obtained by applying $L^{\text{dec}}$ MHA blocks over the partial solution:
\begin{align}
    H_{t,0}&=[h_{1,0}\ \cdots\ h_{t,0}]^T,\\
	h_{t,l}&=\operatorname{MHA}\left(h_{t,l-1},H_{t,l-1},H_{t,l-1},J_{1,t-1}\right),
\end{align}
where $l\in\{1,\ldots L^{\text{dec}}\}$ is the encoder decoder index and $J_{1,t-1}$ is a matrix of ones.

The probability distribution of actions in state $S_t$ is finally computed by $h^p_{t,0}=h_{t,L^{\text{dec}}}$ through $L^{\text{dec}}$ encoder-decoder MHA blocks and a single-head attention with only query and key:
\begin{align}
    h^p_{t,l}&=\operatorname{MHA}(h^p_{t,l-1}, H^e, H^e, M_t),\\
	\pi_{\boldsymbol{\theta}}(A_t\mid G, S_t)&=\operatorname{softmax}\Big(C\operatorname{tanh}\Big(M_t\odot\big(h^p_{t,L^{\text{dec}}}W_1+\boldsymbol{1}b_1\big)(H^eW_2+\boldsymbol{1}b_2)^T\slash\sqrt{d}\Big)\Big),
\end{align}
where $l\in\{1,\ldots L^{\text{dec}}\}$, $C=10$ is a clipping threshold, $M_t\in\{0,1\}^{v(G)\times v(G)}$ is determined jointly by $\mathcal{A}_{S_t}$ and the problem related constraints, and $W_1,W_2\in\mathbb{R}^{d_e\times d_e}$ and $b_1,b_2\in\mathbb{R}^{1\times d_e}$ are trainable parameters. Specifically, the mask $M_t$ is determined by four masks: edge mask $M^e_t$, unvisited vertex mask $M^u_t$, visited vertex mask $M^v_t$ and constraint related mask $M^c_t$ through Hadamard product, which guarantee that the state has no loop, is a tree, and satisfies the problem constraints. The masks are asymmetric and are generated by treating actions as directed edges, which reflects the form of the reduced action space $\mathcal{A}_s$. For example, the unvisited vertex mask in Figure \ref{fig:neuroprim} indicates that the end of selected action must be a vertex that is not in the current state, i.e., $v_2$, which corresponds to the second column of the mask. The specified form of the masks for the DCMST ($d=2$) problem is given in Table \ref{tab:mask}. The MRCST and STP problems share the first three masks with DCMST, while $M^c_t=I$ for any time step $t$, since these two problems do not have any constraints on the selection of edges.

\begin{table}[!htbp]
	\centering
	\caption{The masks for the DCMST ($d=2$) problem, where $J$ and $O$ represent matrices of all ones and zeros respectively, $I$ is the identity matrix, and $e_{ij}$ denotes the matrix with $1$ in the $i$th row and $j$th column row only and all other matrices are $0$. The mask at time step $t$ is obtained by calculating $M_t=M^e_t\odot M^u_t\odot M^v_t\odot M^c_t$. All these matrices are square matrices of order $v(G)$.}
	\label{tab:mask}
	\begin{tabular}{lcccc}
	\toprule
	Time Step & $M^e_t$ & $M^u_t$ & $M^v_t$ & $M^c_t$ \\
	\midrule
	$t=0$ & $J-I$ & $J$ & $O$ & $J$  \\
	$t=1$ & $J-I$ & $J-\sum_i{e_{iv_0}}$ & $O+\sum_j{e_{v_{0}j}}$ & $J$  \\
	$t\in\{2,\ldots,T\}$ & $J-I$ & $M^u_{t-1}-\sum_i{e_{iv_{t-1}}}$ & $M^u_{t-1}+\sum_j{e_{v_{t-1}j}}$ & $\left.\begin{smallmatrix}\max(O,M^c_{t-1}\\-\sum_i{e_{iu_{t-1}}}-\sum_j{e_{u_{t-1}j}})\end{smallmatrix}\right.$ \\
	\bottomrule
	\end{tabular}
\end{table}

\subsection{Policy optimization}

After sampling $N$ solutions with $\pi_{\boldsymbol{\theta}}$, we obtain a set of trajectories $\{(S_{t-1}^n,A_{t-1}^n,R_t^n)_{t=1}^{T_n}\}_{n=1}^N$, where $T_n$ is the final time step of the $n$th trajectory. The gradient of the expected flat partial return $J(\boldsymbol{\theta})$ can be approximated by the policy gradient theorem \cite{suttonPolicyGradientMethods1999} as follows
\begin{equation}
	\nabla J(\boldsymbol{\theta})\approx \frac{1}{N}\sum_{n=1}^{N}\left(R_{T_n}^n-b\right)\nabla\log\prod_{t=1}^{T_n-1} \pi_{\boldsymbol{\theta}}(A_t^n\mid G,S_t^n),
\end{equation}
where $b=\frac{1}{N}\sum_{n=1}^N R_{T_n,n}$ is the POMO baseline sampled from $N$ trajectories decoded from randomly selected vertices and the start time step $t$ is set to $1$ because the first action is a randomly selected vertex and therefore not considered in the policy. We use Adam \cite{kingmaAdamMethodStochastic2017} as the optimizer. The training procedure is described in Algorithm \ref{alg:reinforce}.

\begin{algorithm}
  \caption{REINFORCE with sampled POMO}\label{alg:reinforce}
  \begin{algorithmic}[1]
    \renewcommand{\algorithmicrequire}{\textbf{Input:}}
    \REQUIRE{number of epochs $E$, batch size $B$, trajectories sampled per instance $N$}
    \STATE{Initialize $\boldsymbol{\theta}$}
    \FOR{$\text{epoch}=1,\ldots,E$}
      \FOR{$i=1,\ldots,B$}
        \STATE{$G_i\leftarrow\operatorname{RandomInstance()}$}
        \FOR{$n=1,\ldots,N$}
          \STATE{$A_0^n\leftarrow\operatorname{RandomSelect}(V(G_i))$}
          \STATE{$\{S_{t-1}^n,A_{t-1}^n,R_t^n\}_{t=2}^{T_n}\leftarrow\operatorname{Rollout}(G_i,\pi_{\boldsymbol{\theta}})$}
        \ENDFOR
      \STATE{$b\leftarrow \frac{1}{N}\sum_{n=1}^N R_{T_n,n}$}
      \STATE{$\nabla J\leftarrow \frac{1}{N}\sum_{n=1}^{N}(R_{T_n,n}-b)\nabla\log(\prod_{t=1}^{T_n-1} \pi_{\boldsymbol{\theta}}(A_{t,n}\mid G_i,S_{t,n}))$}
      \ENDFOR
      \STATE{$\boldsymbol{\theta}\leftarrow\operatorname{Adam}(\boldsymbol{\theta}, \nabla J)$}
    \ENDFOR
  \end{algorithmic}
\end{algorithm}

\section{Experiments}

We concentrate our study on three variations of the spanning tree problem, specifically the DCMST ($d=2$), MRCST, and STP problems. Among these, we focus solely on the DCMST problem with a degree constraint of $2$ due to its perceived difficulty in the literature. For each problem, we modify the masks and rewards and train the model on instances of size 50 generated from $U_{[0,1]\times[0,1]}$ following the methodology in \cite{koolAttentionLearnSolve2019}. Both training and testing are executed on a GeForce RTX 3090 GPU. We determine the gap by utilizing the formula $(\text{Obj.} / \text{BKS} - 1) \times 100\%$, where $\text{Obj.}$ signifies the objective obtained by a specific algorithm and $\text{BKS}$ represents the best known solution. Our code in Python is publicly available.\footnote{\url{https://github.com/CarlossShi/neuroprim}}

\paragraph{Hyperparameters}

We set the dimensions of vector, edge features and hidden layer of FFN to $d_v=256$, $d_e=16$ and $d_f=512$, respectively. We use $H=8$ heads, $L^{\text{enc}}=3$ layers in the encoder and $L^{\text{dec}}=1$ in the decoder. The models are trained for $E=100$ epochs, each with $B=2500$ batches of approximately $512$ instances generated by random seed $0$, where the exact number of instances satisfies a product of graph size $v(G)$ with the number of trajectories $N$ sampled per graph. We use a fixed learning rate $\eta=10^{-4}$.

\paragraph{Decoding strategy}

We present two decoding strategies denoted as NeuroPrim (g) and NeuroPrim (s), where the former generates the output greedily and the latter uses sampling based on the masked attention weights to produce the result.

\paragraph{Run times}

For the DCMST ($d=2$), MRCST and STP problems, the training time for NeuroPrim is 27 m, 47 m and 14 m respectively. The longer time for training the MRCST model is due to the inability to parallelize the reward computation, which is attributed to the specific optimization objective of the problem. Unlike these two problems, the final time step $T$ may differ in STP because the number of terminals varies from instance to instance. Thus, for each batch of training data, we sample the number of terminals from $U(2,v(G))$, and this is why its training time is about half that of the DCMST problem. For testing, we find that the time required for NeuroPrim to solve these problems is linear in $T$. A better solution can be obtained by randomly selecting the initial vertices for decoding and sampling multiple solutions and taking the best.

\subsection{Results on benchmark datasets}

We report the results of benchmark tests on Euclidean space for each problem using our algorithm. Prior to testing, we preprocessed the data by scaling and translating all coordinates to fit within the range of $[0,1]^2$. For the NeuroPrim(g) decoding method, we determined the best outcome by traversing $|\mathcal{A}_{K_0}|$ initial vertices and considered it to be a deterministic algorithm. In the case of the NeuroPrim(s) decoding approach, we assessed the cost by sampling $20|\mathcal{A}_{K_0}|$ initial vertices from the action distribution generated by the policy, and we report the mean, standard deviation and best result.

\paragraph{Degree-constrained Minimum Spanning Tree problem}

We present the computational results on CRD\footnote{\url{https://h3turing.cs.hbg.psu.edu/~bui/data/bui-zrncic-gecco2006.html}} dataset, which contains 10 Euclidean instances each of sizes 30, 50, 70 and 100. By reducing this problem to a TSP problem, we obtained BKS and approximate solutions using the Concorde\footnote{\url{https://www.math.uwaterloo.ca/tsp/concorde.html}} and farthest insertion algorithm\footnote{\url{https://github.com/wouterkool/attention-learn-to-route/blob/master/problems/tsp/tsp\_baseline.py}}, respectively. We take the farthest insertion algorithm into account because it is an effective constructive algorithm that adds one vertex per step and is therefore very similar to the execution of NeuroPrim.

As presented in Table \ref{tab:dcmst}, we find that NeuroPrim outperforms the furthest insertion algorithm in terms of optimality gap on all instances. If enough trajectories are sampled, NeuroPrim (s) is able to find optimal solutions for some small-scale problems.

\begin{longtable}[!htbp]{lrr|rr|rrr|rrrr}
\caption{Results for DCMST ($d=2$) problem on CRD instances.}
\label{tab:dcmst}\\
\toprule
 & \multicolumn{2}{c}{} & \multicolumn{2}{c}{Farthest Insertion} & \multicolumn{3}{c}{NeuroPrim (g)} & \multicolumn{4}{c}{NeuroPrim (s)} \\
Ins. & $v(G)$ & BKS & Obj. & Gap & Obj. & Gap & Time & Mean Obj. & Obj. & Gap & Time \\
\midrule
crd300 & 30 & 3822 & 4231 & 10.69\% & 3834 & 0.31\% & 0.2s & 3968$\pm$164 & 3822 & 0.00\% & 2.4s \\
crd301 & 30 & 3616 & 3764 & 4.08\% & 3663 & 1.28\% & 0.2s & 3759$\pm$71 & 3622 & 0.14\% & 2.3s \\
crd302 & 30 & 4221 & 4450 & 5.42\% & 4256 & 0.83\% & 0.1s & 4375$\pm$104 & 4221 & 0.00\% & 2.2s \\
crd303 & 30 & 4233 & 4440 & 4.88\% & 4427 & 4.56\% & 0.1s & 4503$\pm$89 & 4356 & 2.89\% & 2.2s \\
crd304 & 30 & 4274 & 4432 & 3.69\% & 4300 & 0.60\% & 0.1s & 4423$\pm$106 & 4300 & 0.60\% & 2.2s \\
crd305 & 30 & 4252 & 4490 & 5.60\% & 4266 & 0.33\% & 0.1s & 4473$\pm$164 & 4266 & 0.33\% & 2.2s \\
crd306 & 30 & 4214 & 4504 & 6.89\% & 4318 & 2.48\% & 0.1s & 4394$\pm$97 & 4266 & 1.23\% & 2.2s \\
crd307 & 30 & 4261 & 4572 & 7.30\% & 4311 & 1.16\% & 0.1s & 4413$\pm$128 & 4281 & 0.48\% & 2.2s \\
crd308 & 30 & 4029 & 4122 & 2.31\% & 4029 & 0.00\% & 0.1s & 4116$\pm$98 & 4029 & 0.00\% & 2.2s \\
crd309 & 30 & 4031 & 4211 & 4.45\% & 4092 & 1.51\% & 0.1s & 4191$\pm$131 & 4031 & 0.00\% & 2.2s \\
crd500 & 50 & 5312 & 5845 & 10.03\% & 5437 & 2.34\% & 0.3s & 5516$\pm$60 & 5382 & 1.31\% & 4.0s \\
crd501 & 50 & 5556 & 6009 & 8.16\% & 5665 & 1.96\% & 0.2s & 5797$\pm$143 & 5614 & 1.05\% & 3.9s \\
crd502 & 50 & 5482 & 5894 & 7.53\% & 5648 & 3.03\% & 0.2s & 5716$\pm$82 & 5538 & 1.03\% & 3.9s \\
crd503 & 50 & 5085 & 5493 & 8.02\% & 5102 & 0.33\% & 0.2s & 5183$\pm$70 & 5086 & 0.02\% & 4.2s \\
crd504 & 50 & 5306 & 5878 & 10.77\% & 5420 & 2.13\% & 0.2s & 5566$\pm$89 & 5359 & 0.98\% & 4.3s \\
crd505 & 50 & 5513 & 6040 & 9.56\% & 5691 & 3.22\% & 0.2s & 5828$\pm$94 & 5582 & 1.24\% & 4.3s \\
crd506 & 50 & 5179 & 5688 & 9.83\% & 5210 & 0.60\% & 0.2s & 5303$\pm$78 & 5204 & 0.48\% & 4.3s \\
crd507 & 50 & 5230 & 5520 & 5.54\% & 5296 & 1.25\% & 0.2s & 5360$\pm$71 & 5265 & 0.66\% & 4.3s \\
crd508 & 50 & 5369 & 5819 & 8.38\% & 5399 & 0.56\% & 0.2s & 5596$\pm$127 & 5369 & 0.00\% & 4.3s \\
crd509 & 50 & 5103 & 5544 & 8.65\% & 5278 & 3.43\% & 0.2s & 5425$\pm$117 & 5117 & 0.28\% & 4.3s \\
crd700 & 70 & 6307 & 6750 & 7.04\% & 6424 & 1.85\% & 0.3s & 6704$\pm$168 & 6372 & 1.04\% & 7.6s \\
crd701 & 70 & 6117 & 6280 & 2.67\% & 6261 & 2.36\% & 0.3s & 6424$\pm$129 & 6141 & 0.38\% & 6.8s \\
crd702 & 70 & 6768 & 7200 & 6.38\% & 6978 & 3.10\% & 0.3s & 7078$\pm$119 & 6793 & 0.36\% & 6.3s \\
crd703 & 70 & 6305 & 6873 & 9.01\% & 6322 & 0.26\% & 0.3s & 6626$\pm$135 & 6322 & 0.26\% & 6.4s \\
crd704 & 70 & 6372 & 6807 & 6.83\% & 6505 & 2.10\% & 0.3s & 6768$\pm$181 & 6436 & 1.01\% & 6.3s \\
crd705 & 70 & 6456 & 7190 & 11.36\% & 6642 & 2.88\% & 0.3s & 6811$\pm$139 & 6530 & 1.15\% & 6.3s \\
crd706 & 70 & 6674 & 7151 & 7.16\% & 6973 & 4.49\% & 0.3s & 7039$\pm$120 & 6738 & 0.97\% & 6.4s \\
crd707 & 70 & 6468 & 6750 & 4.37\% & 6730 & 4.05\% & 0.3s & 6942$\pm$129 & 6628 & 2.48\% & 6.4s \\
crd708 & 70 & 6233 & 6613 & 6.10\% & 6274 & 0.66\% & 0.3s & 6566$\pm$142 & 6257 & 0.39\% & 6.3s \\
crd709 & 70 & 6046 & 6739 & 11.46\% & 6227 & 2.98\% & 0.3s & 6382$\pm$117 & 6173 & 2.10\% & 6.4s \\
crd100 & 100 & 7049 & 7226 & 2.52\% & 7215 & 2.36\% & 0.6s & 7560$\pm$153 & 7173 & 1.77\% & 12.9s \\
crd101 & 100 & 7697 & 8234 & 6.98\% & 7964 & 3.47\% & 0.6s & 8318$\pm$200 & 7857 & 2.08\% & 12.2s \\
crd102 & 100 & 7534 & 8249 & 9.48\% & 7737 & 2.70\% & 0.6s & 8095$\pm$164 & 7748 & 2.83\% & 12.2s \\
crd103 & 100 & 7666 & 8484 & 10.68\% & 7867 & 2.62\% & 0.6s & 8225$\pm$174 & 7809 & 1.87\% & 12.2s \\
crd104 & 100 & 7539 & 8106 & 7.52\% & 7831 & 3.88\% & 0.6s & 8138$\pm$177 & 7736 & 2.62\% & 12.2s \\
crd105 & 100 & 7080 & 7662 & 8.21\% & 7323 & 3.43\% & 0.6s & 7705$\pm$213 & 7213 & 1.87\% & 12.2s \\
crd106 & 100 & 6883 & 7625 & 10.78\% & 7204 & 4.66\% & 0.6s & 7443$\pm$176 & 7046 & 2.37\% & 12.2s \\
crd107 & 100 & 7724 & 8670 & 12.25\% & 8087 & 4.70\% & 0.6s & 8457$\pm$202 & 7989 & 3.43\% & 12.1s \\
crd108 & 100 & 7136 & 8071 & 13.11\% & 7235 & 1.39\% & 0.6s & 7618$\pm$179 & 7197 & 0.86\% & 12.2s \\
crd109 & 100 & 7290 & 7895 & 8.30\% & 7558 & 3.67\% & 0.6s & 7816$\pm$150 & 7427 & 1.88\% & 12.2s \\
\midrule
Mean &  &  &  & 7.60\% &  & 2.24\% &  &  &  & 1.11\% &  \\
\bottomrule
\end{longtable}

\paragraph{Minimum Routing Cost Spanning Tree problem}

We consider 21 instances of sizes 50, 100, 250 obtained from OR-Library\footnote{\url{http://people.brunel.ac.uk/~mastjjb/jeb/orlib/esteininfo.html}}, which were originally benchmark for Euclidean Steiner problem, but were first used by \cite{julstromBlobCodeCompetitive2005} for testing on the MRCST problem. Since solving MRCST exactly is extremely expensive (e.g., \cite{zetinaSolvingOptimumCommunication2019} took a day to reduce the gap of a 100-size instance to about 10\%), we use the experimental results reported in \cite{sattariMetaheuristicAlgorithmMinimum2015} as BKS. We implement the heuristic proposed by Campos \cite{camposFastAlgorithmComputing2008} using Python and report the results. Campos's algorithm is chosen because it is essentially a variant of Prim's algorithm, with the difference that the metric for selecting edges is changed.

The results are shown in Table \ref{tab:mrcst}. We found that NeuroPrim significantly outperforms Campos's algorithm for instances of size 100 or less. For problems as large as 250, sampling is effective in reducing the optimization gap of the solution.

\begin{table}[!htbp]
\centering
\caption{Results for MRCST problem on OR-Lib instances.}
\label{tab:mrcst}
\begin{tabular}{lrr|rr|rrr|rrrr}
\toprule
 & \multicolumn{2}{c}{} & \multicolumn{2}{c}{Campos} & \multicolumn{3}{c}{NeuroPrim (g)} & \multicolumn{4}{c}{NeuroPrim (s)} \\
Ins. & $v(G)$ & BKS & Obj. & Gap & Obj. & Gap & Time & Mean Obj. & Obj. & Gap & Time \\
\midrule
e50.1 & 50 & 984 & 1028 & 4.48\% & 987 & 0.34\% & 0.4s & 988$\pm$1 & 987 & 0.31\% & 6.9s \\
e50.2 & 50 & 901 & 953 & 5.78\% & 914 & 1.43\% & 0.2s & 915$\pm$1 & 914 & 1.42\% & 5.2s \\
e50.3 & 50 & 888 & 935 & 5.27\% & 892 & 0.44\% & 0.2s & 893$\pm$1 & 892 & 0.37\% & 5.2s \\
e50.4 & 50 & 777 & 846 & 8.96\% & 786 & 1.16\% & 0.2s & 787$\pm$1 & 784 & 0.96\% & 5.2s \\
e50.5 & 50 & 848 & 910 & 7.35\% & 850 & 0.26\% & 0.2s & 851$\pm$0 & 850 & 0.24\% & 5.3s \\
e50.6 & 50 & 818 & 869 & 6.22\% & 834 & 1.96\% & 0.2s & 833$\pm$1 & 831 & 1.57\% & 5.3s \\
e50.7 & 50 & 866 & 907 & 4.79\% & 882 & 1.91\% & 0.2s & 883$\pm$1 & 879 & 1.53\% & 5.2s \\
e100.1 & 100 & 3507 & 3702 & 5.57\% & 3551 & 1.26\% & 0.8s & 3559$\pm$6 & 3542 & 1.01\% & 19.7s \\
e100.2 & 100 & 3308 & 3497 & 5.70\% & 3348 & 1.21\% & 0.8s & 3355$\pm$7 & 3341 & 1.01\% & 16.8s \\
e100.3 & 100 & 3566 & 3736 & 4.76\% & 3624 & 1.61\% & 0.8s & 3625$\pm$4 & 3608 & 1.16\% & 16.3s \\
e100.4 & 100 & 3448 & 3643 & 5.64\% & 3504 & 1.63\% & 0.8s & 3511$\pm$8 & 3490 & 1.22\% & 16.4s \\
e100.5 & 100 & 3637 & 3857 & 6.05\% & 3761 & 3.41\% & 0.8s & 3769$\pm$5 & 3753 & 3.18\% & 16.3s \\
e100.6 & 100 & 3436 & 3601 & 4.78\% & 3480 & 1.26\% & 0.8s & 3489$\pm$6 & 3477 & 1.18\% & 16.2s \\
e100.7 & 100 & 3704 & 3914 & 5.69\% & 3739 & 0.97\% & 0.8s & 3747$\pm$6 & 3737 & 0.92\% & 16.2s \\
e250.1 & 250 & 22088 & 23324 & 5.60\% & 22758 & 3.03\% & 15.2s & 22905$\pm$182 & 22651 & 2.55\% & 5.6m \\
e250.2 & 250 & 22771 & 24031 & 5.53\% & 23307 & 2.35\% & 14.9s & 23379$\pm$37 & 23249 & 2.10\% & 5.5m \\
e250.3 & 250 & 21871 & 23241 & 6.26\% & 22298 & 1.95\% & 14.7s & 22415$\pm$59 & 22253 & 1.74\% & 5.3m \\
e250.4 & 250 & 23423 & 24660 & 5.28\% & 23899 & 2.03\% & 15.0s & 24019$\pm$89 & 23788 & 1.56\% & 6.0m \\
e250.5 & 250 & 22378 & 23517 & 5.09\% & 22915 & 2.40\% & 14.7s & 23075$\pm$80 & 22855 & 2.13\% & 5.7m \\
e250.6 & 250 & 22285 & 23282 & 4.47\% & 22619 & 1.50\% & 14.9s & 22790$\pm$80 & 22584 & 1.34\% & 5.8m \\
e250.7 & 250 & 22909 & 24017 & 4.84\% & 23477 & 2.48\% & 14.9s & 23533$\pm$57 & 23367 & 2.00\% & 5.8m \\
\midrule
Mean &  &  &  & 5.62\% &  & 1.65\% &  &  &  & 1.40\% &  \\
\bottomrule
\end{tabular}
\end{table}

\paragraph{Steiner Tree Problem in graphs}

We use the benchmark P4E\footnote{\url{http://steinlib.zib.de/showset.php?P4E}} instances. These instances consist of complete graph of sizes 100 and 200 with Euclidean weights. The BKS for the instances are given together with the data. We obtain the approximate solution by simply applying the Kruskal's algorithm implemented by SciPy\footnote{\url{https://github.com/scipy/scipy}} on the induced subgraph of all the terminals, as there is little difference in the performance of the algorithms on this problem.

Based on the result presented in Table \ref{tab:stp}, it is evident that NeuroPrim struggles to effectively learn the process of identifying Steiner points to narrow the optimization gap. The performance gap may be attributed to the nature of the problem and the under-sampling of instances that exceed the size of the training set when the stochastic algorithm is executed. Despite this limitation, the algorithm's overall performance is comparable to that of Kruskal's algorithm.

\begin{table}[!htbp]
\centering
\caption{Results for STP on P4E instances.}
\label{tab:stp}
\begin{tabular}{lrrr|rr|rrr|rrrr}
\toprule
 & \multicolumn{3}{c}{} & \multicolumn{2}{c}{Kruskal} & \multicolumn{3}{c}{NeuroPrim (g)} & \multicolumn{4}{c}{NeuroPrim (s)} \\
Ins. & $v(G)$ & $|V^\prime|$ & BKS & Obj. & Gap & Obj. & Gap & Time & Mean Obj & Obj. & Gap & Time \\
\midrule
p455 & 100 & 5 & 1138 & 1166 & 2.46\% & 1166 & 2.46\% & 0.1s & 1166$\pm$0 & 1166 & 2.46\% & 0.7s \\
p456 & 100 & 5 & 1228 & 1239 & 0.90\% & 1239 & 0.90\% & 0.0s & 1239$\pm$0 & 1239 & 0.90\% & 0.5s \\
p457 & 100 & 10 & 1609 & 1642 & 2.05\% & 1642 & 2.05\% & 0.1s & 1642$\pm$1 & 1642 & 2.05\% & 1.0s \\
p458 & 100 & 10 & 1868 & 1868 & 0.00\% & 1868 & 0.00\% & 0.1s & 1884$\pm$17 & 1868 & 0.00\% & 0.9s \\
p459 & 100 & 20 & 2345 & 2348 & 0.13\% & 2348 & 0.13\% & 0.4s & 2365$\pm$13 & 2348 & 0.13\% & 2.1s \\
p460 & 100 & 20 & 2959 & 3010 & 1.72\% & 3015 & 1.89\% & 0.1s & 3030$\pm$14 & 3010 & 1.72\% & 1.8s \\
p461 & 100 & 50 & 4474 & 4492 & 0.40\% & 4499 & 0.56\% & 0.2s & 4528$\pm$17 & 4493 & 0.42\% & 5.1s \\
p463 & 200 & 10 & 1510 & 1545 & 2.32\% & 1545 & 2.32\% & 0.1s & 1545$\pm$0 & 1545 & 2.32\% & 1.4s \\
p464 & 200 & 20 & 2545 & 2574 & 1.14\% & 2574 & 1.14\% & 0.1s & 2593$\pm$13 & 2574 & 1.14\% & 3.4s \\
p465 & 200 & 40 & 3853 & 3880 & 0.70\% & 3880 & 0.70\% & 0.4s & 3972$\pm$504 & 3880 & 0.70\% & 9.7s \\
p466 & 200 & 100 & 6234 & 6253 & 0.30\% & 6285 & 0.82\% & 1.7s & 6527$\pm$143 & 6341 & 1.72\% & 35.0s \\
\midrule
Mean &  &  &  &  & 1.10\% &  & 1.18\% &  &  &  & 1.23\% &  \\
\bottomrule
\end{tabular}
\end{table}

\subsection{Generalization on larger instances}

We further test NeuroPrim for generalizability at problem sizes on instances as large as 1000. We randomly generate test instances of size $n=50, 100, 200, 500, 1000$ on $U_{[0,1],[0,1]}$, with $10000/n$ of each kind, sample 10 results of greedy decoding each, and average the best results obtained by instance size. The BKS for the DCMST ($d=2$) and STP problems are obtained by the Concorde solver and the exact solving framework proposed by Leitner \cite{leitnerDualAscentBasedBranchandBound2018}, respectively. For the MRCST problem, we use the Campos's algorithm to obtain the BKS due to the difficulty of solving the exact solutions. From the results in Table \ref{tab:gen}, we can see that the optimality gap obtained by NeuroPrim increases gradually with the problem size. We obtain gaps of 13.44\% and 1.44\% for the DCMST ($d=2$) and STP problems of size 1000, respectively, while the MRCST problem outperforms the heuristic results in all tested sizes.

\begin{table}[!htbp]
	\centering
	\caption{Generalization ability of NeuroPrim on random instances.}
	\label{tab:gen}
	\begin{tabular}{lrrrrr}
	\toprule
	Problem & 50 & 100 & 200 & 500 & 1000 \\
	\midrule
	DCMST ($d=2$) & 1.67\% & 3.89\% & 7.80\% & 11.74\% & 13.44\% \\
	MRCST & -4.86\% & -4.08\% & -3.20\% & -1.90\% & -1.27\% \\
	STP ($|V^\prime|=v(G)/2$) & 0.82\% & 0.91\% & 1.08\% & 1.39\% & 1.44\% \\
	\bottomrule
	\end{tabular}
\end{table}

\subsection{Sensitivity Analysis}

\begin{figure}[!htbp]
	\centering
    \includegraphics[width=\textwidth, angle=0]{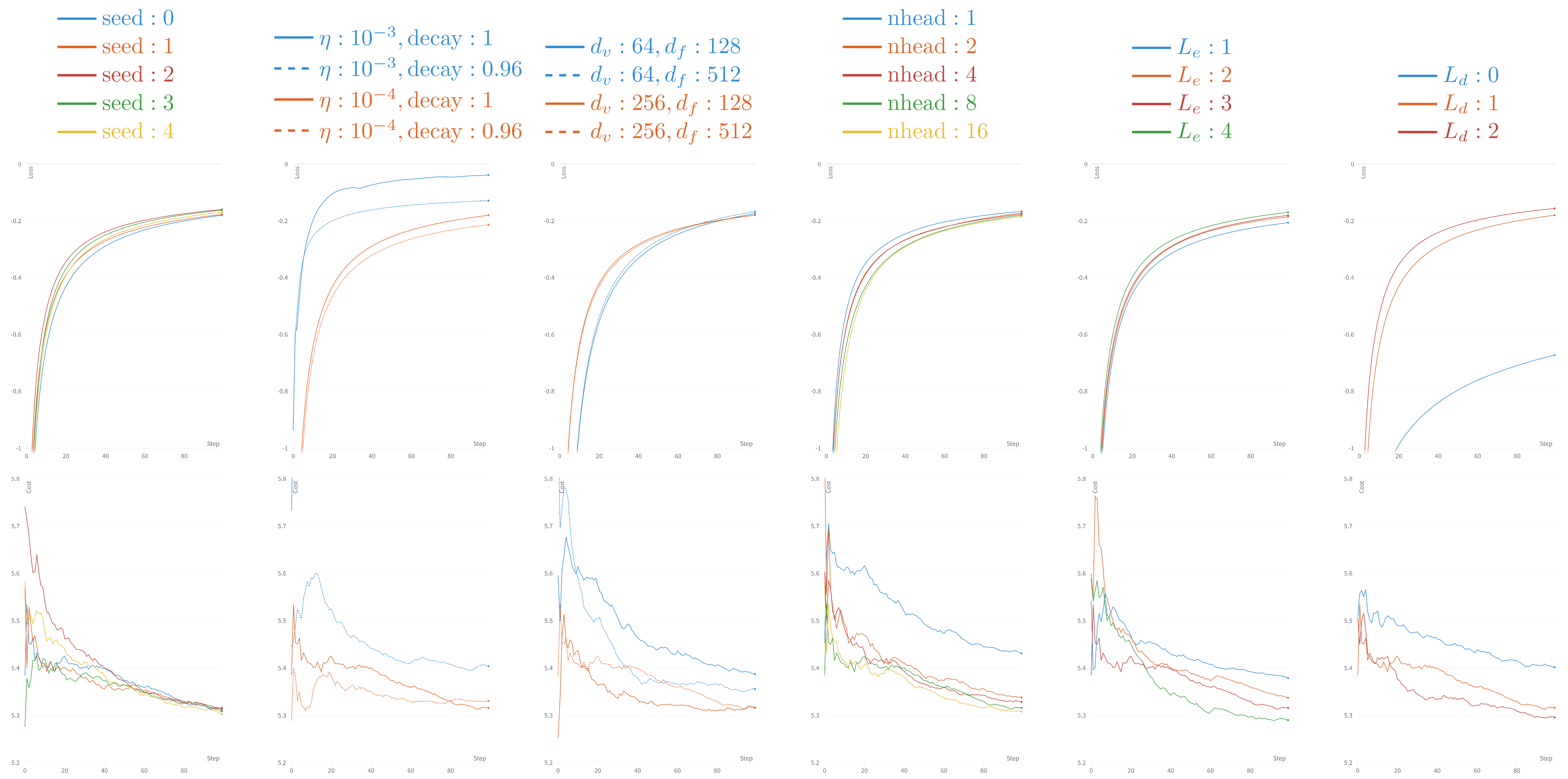}
	\caption{Sensitivity analysis of hyperparameters on the DCMST ($d=2$) problem.}
	\label{fig:sens_dcmst}
\end{figure}

\begin{figure}[!htbp]
	\centering
    \includegraphics[width=\textwidth, angle=0]{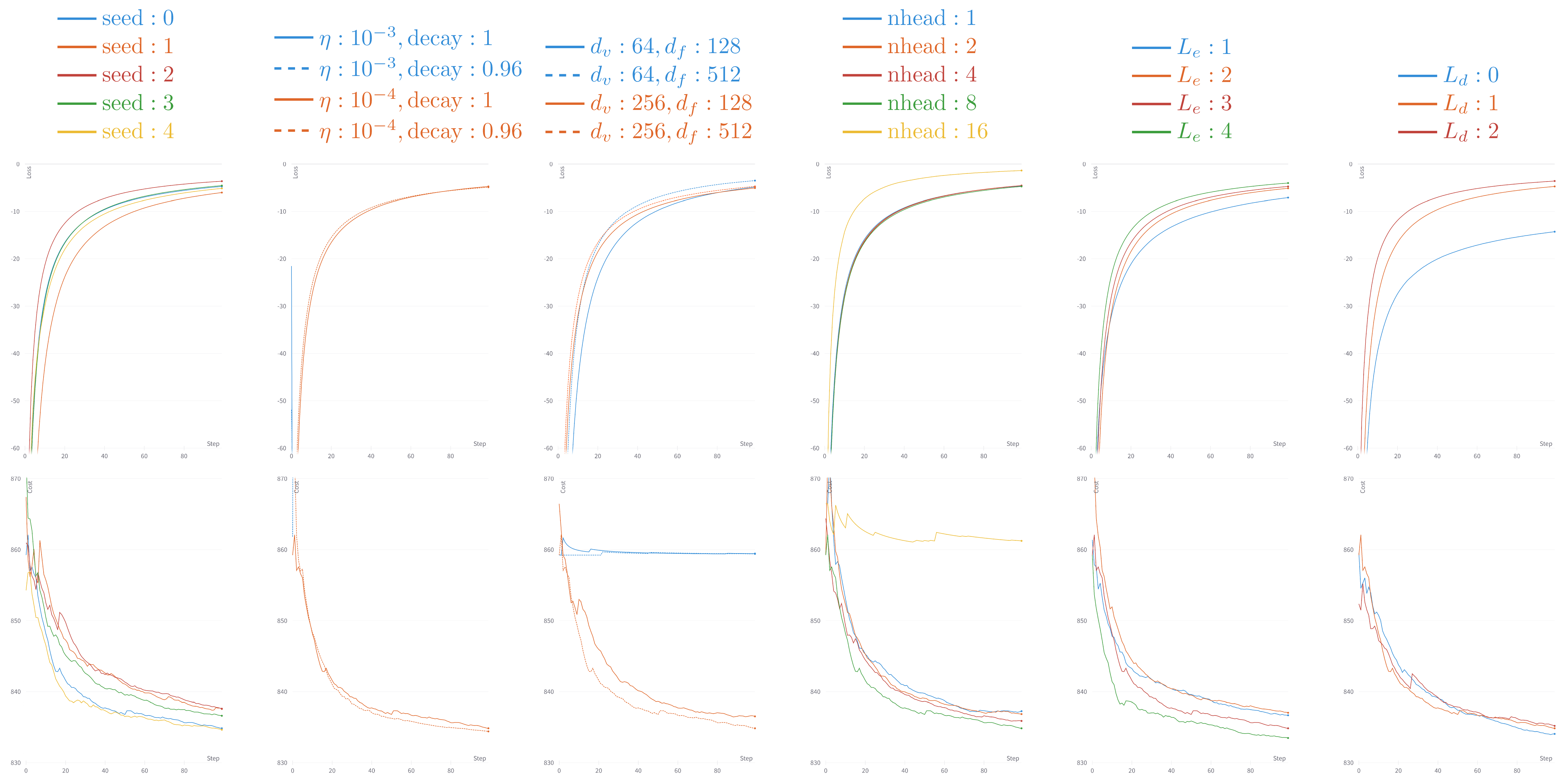}
	\caption{Sensitivity analysis of hyperparameters on the MRCST problem.}
	\label{fig:sens_mrcst}
\end{figure}

\begin{figure}[!htbp]
	\centering
    \includegraphics[width=\textwidth, angle=0]{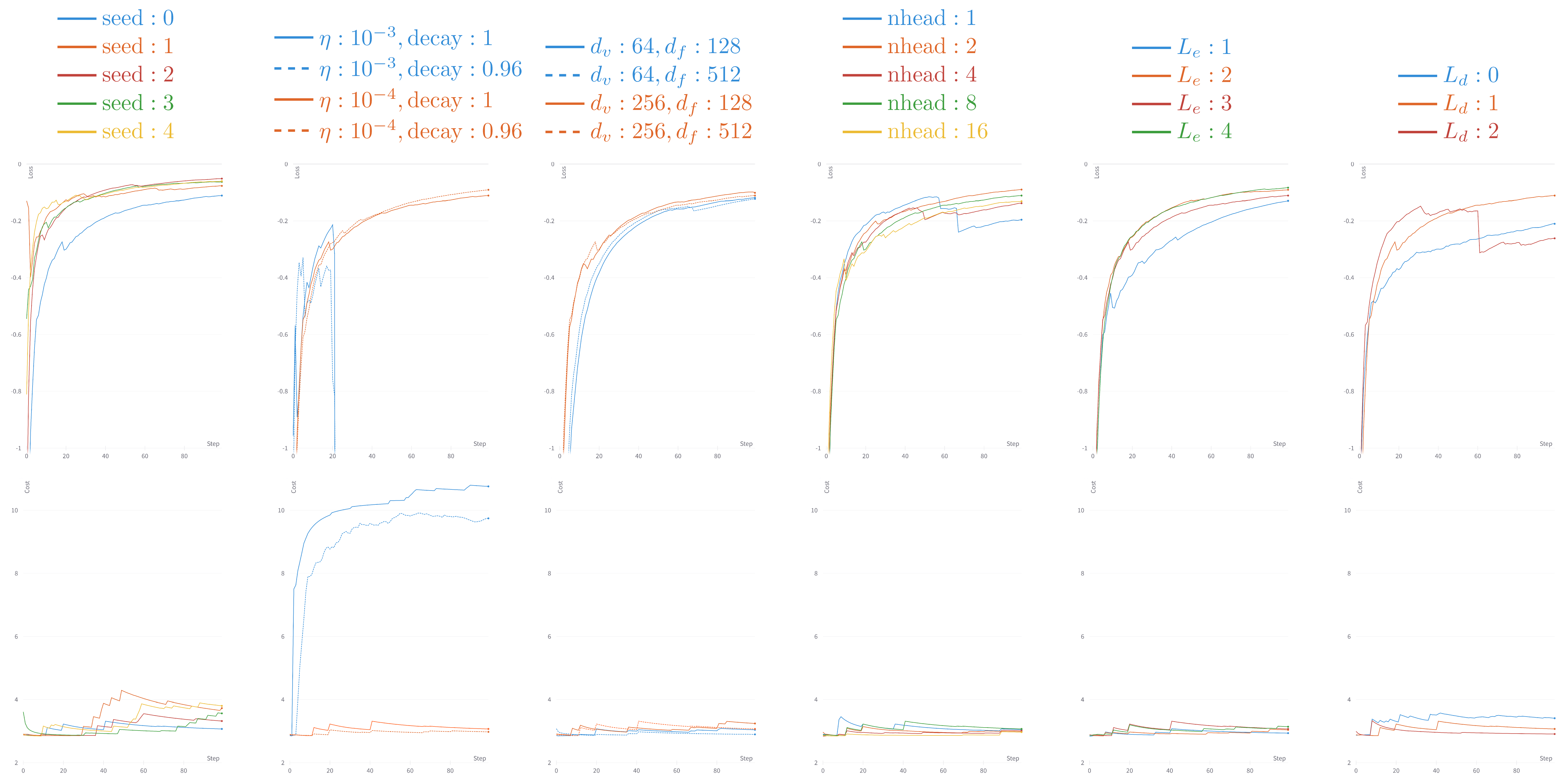}
	\caption{Sensitivity analysis of hyperparameters on STP}
	\label{fig:sens_stp}
\end{figure}

This section presents a one-at-a-time sensitivity analysis of six hyperparameters: random seed, learning rate $\eta$, model dimension $d_e$ and $d_f$, number of heads used in MHA blocks, number of encoder layers and number of decoder layers. The remaining parameters are kept constant as specified above, except for the batch size $B$, which is set to $1000$. Specifically, each hyperparameter is individually adjusted while holding the remaining parameters constant. The aim is to observe the resulting training losses and performance on randomly generated test instances for each problem. Figures \ref{fig:sens_dcmst}, \ref{fig:sens_mrcst}, and \ref{fig:sens_stp} illustrate the loss and cost results obtained from training in DCMST ($d=2$), MRCST and STP problems, respectively, plotted against the model epoch.

Our findings indicate that the learning rate is a crucial hyperparameter for achieving model convergence and optimal performance in all problems. A large learning rate can lead to a failure to converge; hence, a straightforward and effective choice is to set the learning rate to $0.0001$. Although increasing the model dimension, the number of heads, and the number of encoding layers typically improves the trained algorithm's performance, the improvement is usually not significant, except in the MRCST problem. Here, setting the number of heads to $16$ may result in a local optimum, and thus, a simpler and more stable option is to set the number of headsd to $8$. Additionally, the number of decoding layers typically enhances the algorithm's performance, especially in the DCMST ($d=2$) problem. Our analysis shows that the choice of random seeds has minimal effect on the effectiveness of model training.

\subsection{Example solutions}

\begin{figure}[!htbp]
	\centering
	\begin{subfigure}{.3\textwidth}
		\includegraphics[width=\textwidth]{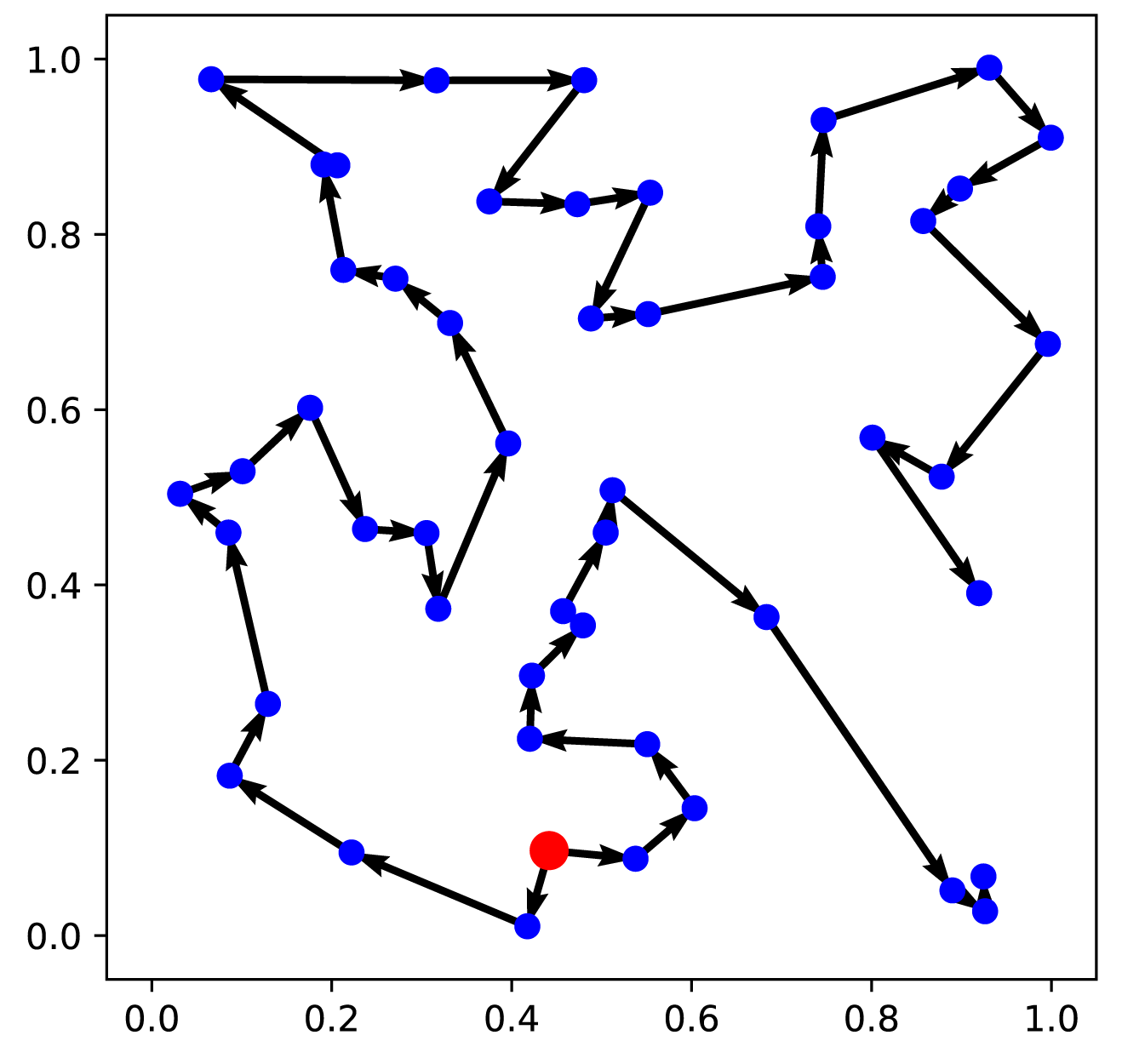}
		\caption{DCMST ($d=2$) of Obj. $6.06$}
	\end{subfigure}
	\begin{subfigure}{.3\textwidth}
		\includegraphics[width=\textwidth]{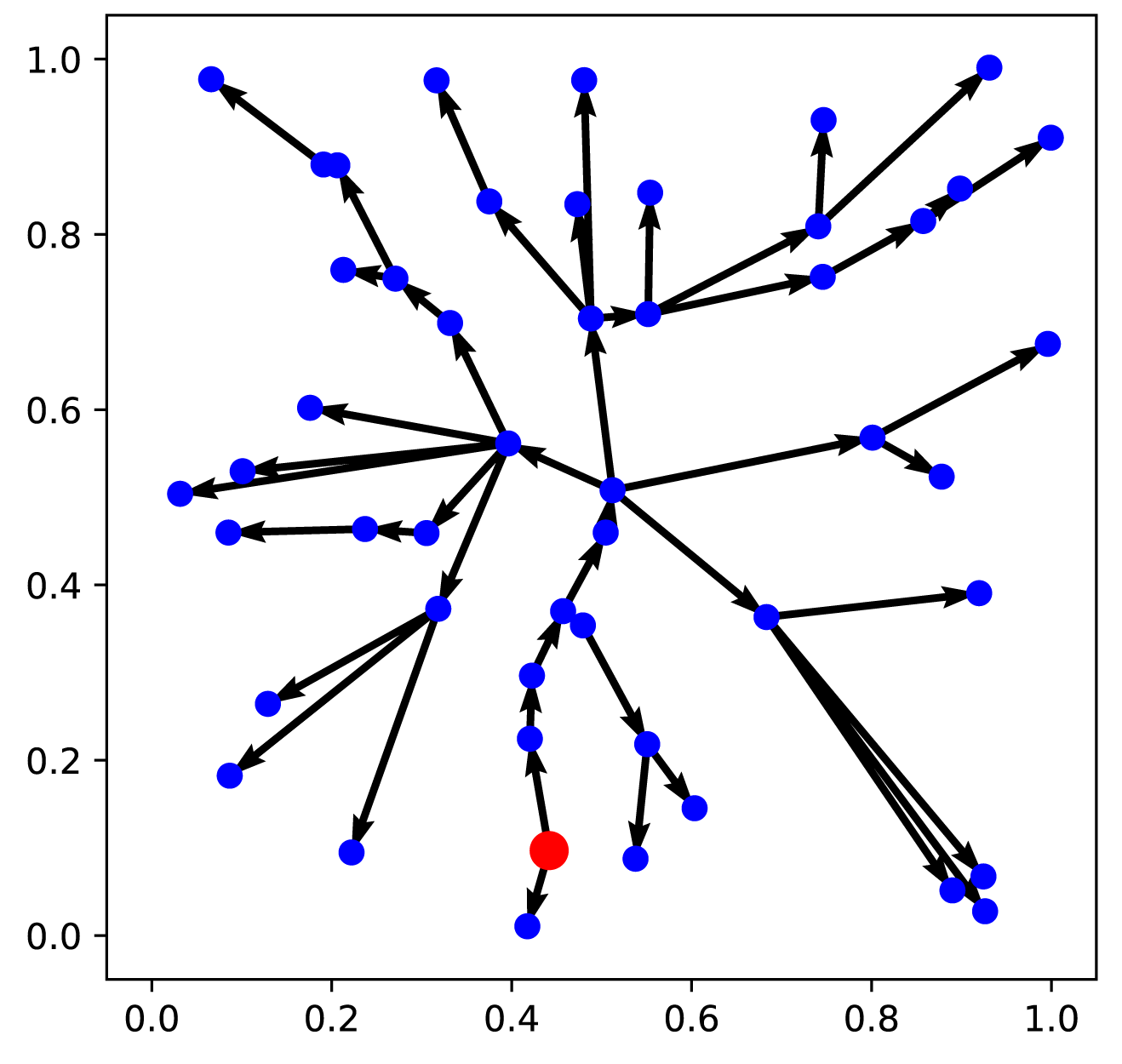}
		\caption{MRCST of Obj. $917.5$}
	\end{subfigure}
	\begin{subfigure}{.3\textwidth}
		\includegraphics[width=\textwidth]{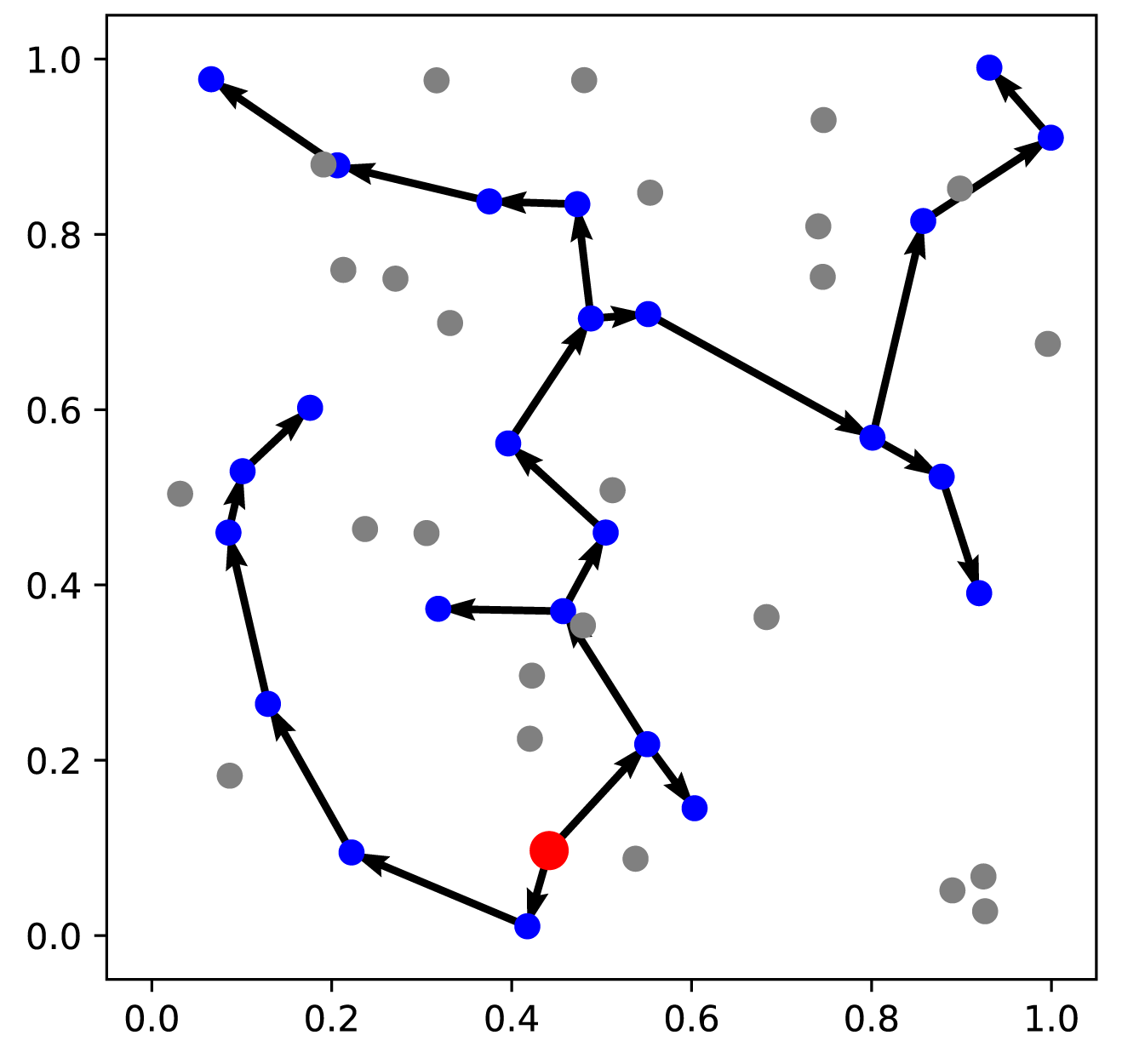}
		\caption{STP ($|V^\prime|=25$) of Obj. $3.54$}
	\end{subfigure}
	\caption{Example solutions of an instance of size 50 for DCMST ($d=2$), MRCST, STP ($|V^\prime|=25$). The terminals are represented by blue nodes, and the edges in solutions by black arrows. The vertex at the beginning of the solution are highlighted in red.}
	\label{fig:visual}
\end{figure}

As shown in Figure \ref{fig:visual}, we choose an instance of size 50 and visualize solutions in three problems, i.e., DCMST ($d=2$), MRCST, STP ($|V^\prime|=25$), using greedy decoding strategy. For DCMST, the edges in the solution set are not interleaved and strictly adhere to the degree constraint of the vertices. In MRCST, the set of edges radiates outwards from the center, which is consistent with the goal of minimizing the average pairwise distance between two vertices. As for STP, the model is able to distinguish between terminals and ordinary vertices, and constructs the solution set step by step according to the distribution of terminals. The results show that NeuroPrim is able to learn new algorithms efficiently for general combinatorial optimization problems on graphs with complex constraints and few heuristics.

\section{Conclusion}

In this paper, we introduced a unified framework called NeuroPrim for solving various spanning tree problems on Euclidean space, by defining the MDP for general combinatorial optimization problems on graphs. Our framework achieves near-optimal results in a short time for the DCMST, MRCST and STP problems, without requiring extensive problem-related knowledge. Moreover, it generalizes well to larger instances. Notably, our approach can be extended to solve a wider range of combinatorial optimization problems beyond spanning tree problems.

Looking ahead, there is a need to explore more general methods for solving combinatorial optimization problems. While there are existing open source solution frameworks such as \cite{hubbsORGymReinforcementLearning2020, zhengOpenGraphGymParallelReinforcement2020}, they often rely on designing specific algorithm components for individual problems. Additionally, scaling up to larger problems poses a challenge, as the attention matrix may become too large to store in memory. A possible solution to this is to use local attention within a neighborhood, rather than global attention, which may unlock further potential for the efficiency of our algorithm.

\section*{Acknowledgements}

This research was supported by National Key R\&D Program of China (2021YFA1000403), the National Natural Science Foundation of China (Nos. 11991022), the Strategic Priority Research Program of Chinese Academy of Sciences (Grant No. XDA27000000) and the Fundamental Research Funds for the Central Universities.

\bibliographystyle{unsrtnat}

\end{document}